\documentclass[9pt,twocolumn,twoside]{osajnl}

\journal{ol} 
\usepackage{units}
\usepackage{rotating}

\newcommand\revise[1]{#1}

\newcommand{\vect}[1]{\boldsymbol{#1}}

\newcommand{\argmax}{\operatornamewithlimits{argmax}}
\setboolean{shortarticle}{true}

\title{\Huge{Learning Agile and Dynamic Motor Skills for Legged Robots}}

\author[1*]{Jemin Hwangbo}
\author[1]{Joonho Lee}
\author[2]{Alexey Dosovitskiy}
\author[1]{Dario Bellicoso}
\author[1]{Joonho Lee}
\author[1]{Vassilios Tsounis}
\author[2]{Vladlen Koltun}
\author[1]{Marco Hutter}

\affil[1]{Robotic Systems Lab, ETH Zurich}
\affil[2]{Intelligent Systems Lab, Intel}
\affil[*]{Corresponding author: jemin.hwangbo@gmail.com}

\begin{abstract}
Legged robots pose one of the greatest challenges in robotics. Dynamic and agile maneuvers of animals cannot be imitated by existing methods that are crafted by humans. A compelling alternative is reinforcement learning, which requires minimal craftsmanship and promotes the natural evolution of a control policy. However, so far, reinforcement learning research for legged robots is mainly limited to simulation, and only few and comparably simple examples have been deployed on real systems. The primary reason is that training with real robots, particularly with dynamically balancing systems, is complicated and expensive. In the present work, we report a new method for training a neural network policy in simulation and transferring it to a state-of-the-art legged system, thereby we leverage fast, automated, and cost-effective data generation schemes. The approach is applied to the ANYmal robot, a sophisticated medium-dog-sized quadrupedal system. Using policies trained in simulation, the quadrupedal machine achieves locomotion skills that go beyond what had been achieved with prior methods: ANYmal is capable of precisely and energy-efficiently following high-level body velocity commands, running faster than ever before, and recovering from falling even in complex configurations.
\end{abstract}

\setboolean{displaycopyright}{false}

\begin{document}

\maketitle

\section*{Introduction}
Legged robotic systems are attractive alternatives to tracked/wheeled robots for applications in rough terrain and complex cluttered environments. Their freedom to choose contact points with the environment enables them to overcome obstacles comparable to their leg length. With such capabilities, legged robots may one day rescue people in forests and mountains, climb stairs to carry payloads in construction sites, inspect unstructured underground tunnels, and explore other planets. Legged systems have the potential to perform any physical activity humans and animals are capable of.

A variety of legged systems are being developed in the effort to take us closer to this vision of the future. Boston Dynamics introduced a series of robots equipped with hydraulic actuators~\cite{raibert2008bigdog,nelson2012petman}. These have advantages in operation since they are powered by conventional fuel with high energy density. However, systems of this type cannot be scaled down (usually $>$ \unit[40]{kg}) and generate smoke and noise, limiting them to outdoor environments. Another family of legged systems is equipped with electric actuators, which are better suited to indoor environments. MIT Cheetah~\cite{seok2013design} is one of the most promising legged systems of this kind. It is a fast, efficient, and powerful quadrupedal robot designed with advanced actuation technology. However, it is a research platform optimized mainly for speed and has not been thoroughly evaluated with respect to battery life, turning capability, mechanical robustness, and outdoor applicability. Boston Dynamics' newly introduced robot, SpotMini, is also driven by electric actuators and is designed for both indoor and outdoor applications. Although the details have not been disclosed, a series of public demonstrations and media releases~\cite{BDSpotMini} are convincing evidence of its applicability to real-world operation. The platform used in this work, ANYmal~\cite{hutter2016anymal}, is another promising quadrupedal robot powered by electric actuators. Its bioinspired actuator design makes it robust against impact while allowing accurate torque measurement at the joints. However, the complicated actuator design increases cost and compromises the power output of the robot.

Designing control algorithms for these hardware platforms remains exceptionally challenging. From the control perspective, these robots are high-dimensional and non-smooth systems with many physical constraints. The contact points change over the course of time and depending on the maneuver being executed and, therefore, cannot be prespecified. Analytical models of the robots are often inaccurate and cause uncertainties in the dynamics. A complex sensor suite and multiple layers of software bring noise and delays to information transfer. Conventional control theories are often insufficient to deal with these problems effectively. Specialized control methods developed to tackle this complex problem typically require a lengthy design process and arduous parameter tuning.

The most popular approach to controlling physical legged systems is modular controller design. This method breaks the control problem down into smaller submodules that are largely decoupled and are therefore easier to manage.
Each module is based on template dynamics~\cite{full1999templates} or heuristics and generates reference values for the next module. For example, some popular approaches~\cite{raibert1981hybrid,pratt2006capture,goswami1997limit,schwind1998spring} use a template-dynamics-based control module that approximates the robot as a point mass with a massless limb to compute the next foothold position. Given the foothold positions, the next module computes a parameterized trajectory for the foot to follow. The last module tracks the trajectory with a simple Proportional-Integral-Derivative (PID) controller.
Since the outputs of these modules are physical quantities, such as body height or foot trajectory, each module can be individually hand-tuned. Approaches of this type have achieved impressive results. Kalakrishnan et al.~\cite{kalakrishnan2010fast} demonstrated robust locomotion over challenging terrain with a quadrupedal robot: to date this remains the state-of-the-art for rough terrain locomotion. Recently, Bellicoso et al.~\cite{bellicoso2018dynamic} demonstrated dynamic gaits, smooth transitions between them, and agile outdoor locomotion with a similar controller design.
Yet despite their attractive properties, modular designs have limitations.
First, limited detail in the modeling constrains the model's accuracy.
This inherent drawback is typically mitigated by limiting the operational state domain of each module to a small region where the approximations are valid. In practice, such constraints lead to significant compromises in performance, such as slow acceleration, fixed upright pose of the body, and limited velocity of the limbs.
Second, the design of modular controllers is extremely laborious.
Highly trained engineers spend months to develop a controller and to arduously hand-tune the control parameters per module for every new robot or even for every new maneuver. For example, running and climbing controllers of this kind can have drastically different architectures and are designed and tuned separately.

More recently, trajectory optimization approaches have been proposed to mitigate the aforementioned problems.
In these methods, the controller is separated into two modules: planning and tracking. The planning module uses rigid-body dynamics and numerical optimization to compute an optimal path that the robot should follow to reach the desired goal. The tracking module is then used to follow the path. In general, trajectory optimization for a complex rigid-body model with many unspecified contact points is beyond the capabilities of current optimization techniques. Therefore, in practice, a series of approximations are employed to reduce complexity. Some methods approximate the contact dynamics to be smooth ~\cite{neunert2017trajectory,mordatch2012discovery}, making the dynamics differentiable. Notably, Neunert et al.~\cite{neunert2017trajectory} demonstrated that such methods can be used to control a physical quadrupedal robot. Other methods~\cite{farshidian2017efficient} prespecify the contact timings and solve for sections of trajectories where the dynamics remain smooth.
A few methods aim to solve this problem with little to no approximation~\cite{posa2014direct,carius2018trajectory}. These methods can discover a gait pattern (i.e., contact sequence) with hard contact models and have demonstrated automatic motion generation for 2D robotic systems but, like any other trajectory optimization approach, the possible contact points are specified a priori.
While more automated than modular designs, the existing optimization methods perform worse than state-of-the-art modular controllers.
The primary issue is that numerical trajectory optimization remains challenging, requires tuning, and in many cases can produce suboptimal solutions.
Besides, optimization has to be performed at execution time on the robot, making these methods computationally expensive. This problem is often solved by reducing precision or running the optimization on a powerful external machine, but both solutions introduce their own limitations.
Furthermore, the system still consists of two independent modules that do not adapt to each other's performance characteristics. This necessitates hand-tuning of the tracker; yet accurately tracking fast motion by an underactuated system with many unexpected contacts is nearly impossible.

Data-driven methods, like reinforcement learning (RL), promise to overcome the limitations of prior model-based approaches by learning effective controllers directly from experience.
The idea of RL is to collect data by trial and error and automatically tune the controller to optimize the given cost (or reward) function representing the task.
This process is fully automated and can optimize the controller end-to-end, from sensor readings to low-level control signals, thereby allowing for highly agile and efficient controllers.
On the downside, RL typically requires prohibitively long interaction with the system to learn complex skills~-- typically weeks or months of real-time execution~\cite{levine2018learning}.
Moreover, over the course of training, the controller may exhibit sudden and chaotic behavior, leading to logistical complications and safety concerns.
Direct application of learning methods to physical legged systems is therefore complicated and has only been demonstrated on relatively simple and stable platforms~\cite{tedrake2004stochastic} or in a limited context~\cite{yosinski2011evolving}.

Due to the difficulties of training on physical systems, most advanced applications of RL to legged locomotion are restricted to simulation. Recent innovations in RL make it possible to train locomotion policies for complex legged models. Levine and Koltun~\cite{levine2014learning} combined learning and trajectory optimization to train a locomotion controller for a simulated 2D walker. Schulman et al.~\cite{schulman2015trust} trained a locomotion policy for a similar 2D walker with an actor-critic method.
More recent work obtained full 3D locomotion policies~\cite{schulman2017proximal,heess2017emergence,peng2017deeploco,xie2018feedback}. In these papers, animated characters achieve remarkable motor skills in simulation.

Given the achievements of reinforcement learning in simulated environments, a natural question is whether these learned policies can be deployed on physical systems.
Unfortunately, such simulation-to-reality transfer is hindered by the reality gap~-- the discrepancy between simulation and the real system in terms of dynamics and perception.
There are two general approaches to bridging the reality gap.
The first is to improve simulation fidelity either analytically or in a data-driven way; the latter is also known as system identification~\cite{lee2018geometric,neunert2017off,bongard2006resilient,nguyen2009model,nguyen2008learning,nikzad1996actuator}.
The second approach is to accept the imperfections of simulation and aim to make the controller robust to variations in system properties, thereby allowing for better transfer.
This robustness can be achieved by randomizing various aspects of the simulation: employing a stochastic policy~\cite{sutton1998reinforcement}, randomizing the dynamics~\cite{mordatch2015ensemble,tan2018sim,peng2017sim,jakobi1995noise}, adding noise to the observations~\cite{dosovitskiy2016learning}, and perturbing the system with random disturbances.
Both approaches lead to improved transfer; however, the former is cumbersome and often impossible, while the latter can compromise the performance of the policy.
Therefore, in practice, both are typically employed in conjunction.
For instance, the recent work of Tan et al.~\cite{tan2018sim} demonstrates successful sim-to-real transfer of locomotion policies on a quadrupedal system called Minitaur via the use of an accurate analytical actuator model and dynamics randomization.
Although achieving impressive results, the method of Tan et al.~\cite{tan2018sim} crucially depends on accurate analytical modeling of the actuators, which is possible for direct-drive actuators (as used in Minitaur), but not for more complex actuators such as servomotors, Series-Elastic Actuators (SEAs), and hydraulic cylinders, which are commonly used in larger legged systems.

In this work, we develop a practical methodology for autonomously learning and transferring agile and dynamic motor skills for complex and large legged systems, such as the ANYmal robot~\cite{hutter2016anymal}.
Compared to the robots used in~\cite{tan2018sim}, ANYmal has a much larger leg length relative to footprint, making it more dynamic, less statically stable, and therefore more difficult to control.
In addition, it features 12 SEAs, which are difficult to control and for which sufficiently accurate analytical models do not exist. Gehring et al.~\cite{gehring2016practice} have attempted analytical modeling of an SEA but, as we will show, their model is insufficiently accurate for training a high-performance locomotion controller.

\begin{figure*}
    \centering
    \includegraphics[width=0.9\textwidth]{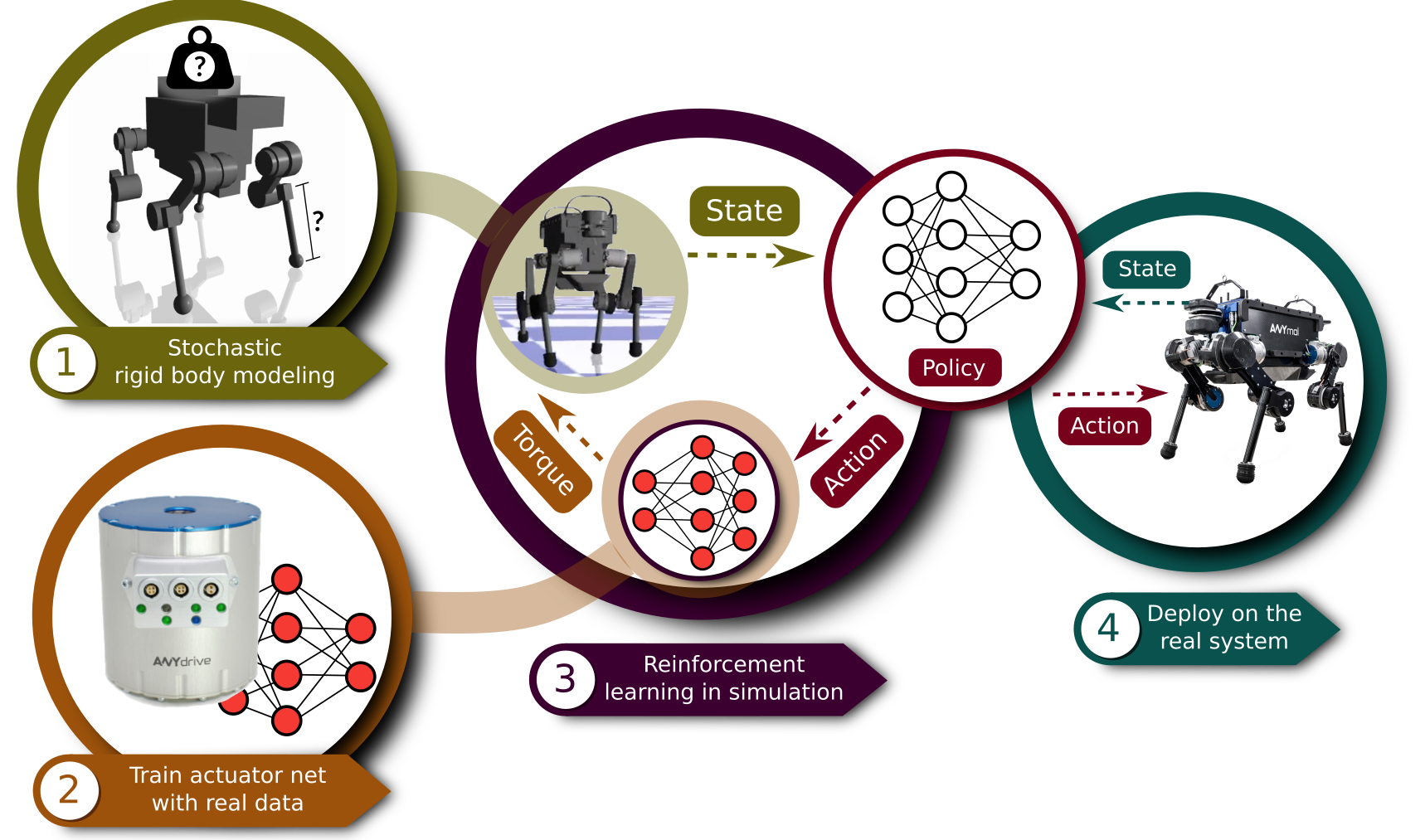}
    \caption{\textbf{Creating a control policy.} In the first step, we identify the physical parameters of the robot and estimate uncertainties in the identification. In the second step, we train an actuator net that models complex actuator/software dynamics. In the third step, we train a control policy using the models produced in the first two steps. In the fourth step, we deploy the trained policy directly on the physical system.}
\end{figure*}

Our approach is summarized in Fig.~1.
Our key insight on the simulation side is that efficiency and realism can be achieved by combining classical models representing well-known articulated system and contact dynamics with learning methods that can handle complex actuation (Fig.~1, steps 1 and 2).
The rigid links of ANYmal, connected through high-quality ball bearings, closely resemble an idealized multi-body system that can be modeled with well-known physical principles \cite{featherstone2014rigid}.
However, this analytical model does not include the set of mechanisms that map the actuator commands to the generalized forces acting on the rigid-body system: the actuator dynamics, the delays in control signals introduced by multiple hardware and software layers, the low-level controller dynamics, and compliance/damping at the joints.
Since these mechanisms are nearly impossible to model accurately, we learn the corresponding mapping in an end-to-end manner~-- from commanded actions to the resulting torques~-- with a deep network.
We learn this ``actuator net'' on the physical system via self-supervised learning and use it in the simulation loop to model each of the 12 joints of ANYmal.
Crucially, the full hybrid simulator, including a rigid-body simulation and the actuator nets, runs at nearly 500K time steps per second, which allows the simulation to run roughly a thousand times faster than real time. About half of the runtime is used to evaluate the actuator nets, and the remaining computations are efficiently performed via 
our in-house simulator, which exploits the fast contact solver of Hwangbo et al.~\cite{hwangbo2018per}, efficient recursive algorithms for computing dynamic properties of articulated systems (composite rigid-body algorithm and recursive Newton-Euler algorithm)~\cite{featherstone2014rigid}, and a fast collision detection library~\cite{smith2005open}. Thanks to efficient software implementations, we did not need any special computing hardware, such as multi-CPU or multi-GPU servers, for training. All training sessions presented in this paper were done on a personal computer with one CPU and one GPU, and none lasted more than eleven hours.

We use the hybrid simulator for training controllers via reinforcement learning (Fig.~1, step 3).
The controller is represented by a multi-layer perceptron that takes as input the history of the robot's states and produces as output the joint position target.
Specifying different reward functions for RL yields controllers for different tasks of interest.

The trained controller is then directly deployed on the physical system (Fig.~1, step 4).
Unlike the existing model-based control approaches, our proposed method is computationally efficient at runtime. Inference of the simple network used in this work takes \unit[$25$]{$\mu s$} on a single CPU thread, which corresponds to about 0.1\% of the available onboard computational resources on the robot used in the experiments. This is in contrast to model-based control approaches that often require an external computer to operate at sufficient frequency~\cite{farshidian2017efficient,neunert2017trajectory}. Also, by simply swapping the network parameter set, the learned controller manifests vastly different behaviors. Although these behaviors are trained separately, they share the same code base: only the high-level task description changes depending on the behavior. In contrast, most of the existing controllers are task-specific and have to be developed nearly from scratch for every new maneuver.

We apply the presented methodology to learning several complex motor skills that are deployed on the physical quadruped. 
First, the controller enables the ANYmal robot to follow base velocity commands more accurately and energy-efficiently than the best previously existing controller running on the same hardware.
Second, the controller makes the robot run faster than ever before, breaking the previous speed record of ANYmal by \unit[25]{\%}.
The controller can operate at the limits of the hardware and push performance to the maximum.
Third, we learn a controller for dynamic recovery from a fall.
This maneuver is exceptionally challenging for existing methods, since it involves multiple unspecified internal and external contacts. It requires fine coordination of actions across all limbs and must use momentum to dynamically flip the robot.
To the best of our knowledge, such recovery skill has never before been achieved on a quadruped of comparable complexity.

\section*{Results}

\href{https://youtu.be/aTDkYFZFWug}{Movie S1} summarizes the results and the method of this work. In the following subsections, we describe the results in detail.

\subsection*{Command-conditioned locomotion}

In most practical scenarios, the motion of a robot is guided by high-level navigation commands, such as the desired direction and speed of motion.
These commands can be provided for instance by an upper-level planning algorithm or by a user via teleoperation.
Using our method, we trained a locomotion policy that can follow such commands at runtime, adapting the gait as needed, with no prior knowledge of command sequence and timing. A command consists of three components: forward velocity, lateral velocity, and yaw rate.

We first qualitatively evaluate this learned locomotion policy by giving random commands using a joystick. Additionally, the robot is disturbed during the experiment by multiple external pushes to the main body.
The resulting behavior is shown in \href{https://youtu.be/23mBeaGmQ2o}{movie S2}.
The video shows about 40 seconds of robust command following. We also tested the policy for five minutes without a single failure, which manifests the robustness of the learned policy.

The trained policy perform stably within the command distribution that it is trained on, with any random combination of the command velocities. Although the forward command velocity is sampled from \unit[$U(-1,1)$]{m/s} during training, the trained policy reaches \unit[1.2]{m/s} of measured forward velocity reliably when the forward command velocity is set slightly higher (\unit[1.23]{m/s}) and the other command velocities are set to zero.

Next, we quantitatively evaluate this learned locomotion policy by driving the robot with randomly-sampled commands. The commands are sampled as described in section S2. The robot receives a new command every two seconds and the command is held constant in between. The test is performed for 30 seconds and a total of 15 random transitions are performed, including the initial transition from zero velocity. The base velocity plot is shown in fig.~S1. The average linear velocity error was \unit[0.143]{m/s} and the average yaw rate error was \unit[0.174]{rad/s}.


We now compare the learned controller to the best-performing existing locomotion controller available for ANYmal~\cite{bellicoso2018dynamic}. For this experiment, we used a flying trot gait pattern (trot with full flight phase), since this is the only gait that stably reaches \unit[1.0]{m/s} forward velocity. We used the same velocity command profile which results in the base velocity shown in fig.~S2. The average linear velocity error was \unit[0.231]{m/s} and the average yaw rate error was \unit[0.278]{rad/s}. Given the same command profile, the tracking error of the model-based controller is about 95\% higher than our learned controller with respect to linear velocity and about 60\% higher with respect to yaw rate. In addition, our learned controller used less torque (\unit[8.23]{Nm} vs. \unit[11.7]{Nm}) and less mechanical power (\unit[78.1]{W} vs. \unit[97.3]{W}) in average. \href{https://youtu.be/aqVPyIgZ15M}{Movie S3} illustrates the experiments for both the learned policy and the model-based policy.

The control performance was also evaluated and compared in forward running. To this end, we sent a step input of four different speed commands (0.25, 0.5, 0.75, and \unit[1.0]{m/s}) for \unit[4.5]{s} each. The results, including a comparison to the prior method~\cite{bellicoso2018dynamic}, are shown in Fig.~2. Figure~2A shows the flying trot pattern discovered by the learned controller. Note that this flight phase disappears for low-velocity commands and ANYmal displays walking trot as shown in movie S2. Even without specifying the gait pattern, the learned policy manifests trot, a gait pattern that is commonly observed in quadrupedal animals. Figure~2B shows the velocity tracking accuracy of the policy both in simulation and on the real robot. Note that the oscillation of the observed velocity around the commanded one is a well-known phenomenon in legged systems, including humans~\cite{winter1991biomechanics}. In terms of average velocity, the learned policy has an error of \unit[2.2]{\%} on the real robot, \unit[1.1]{\%} higher than in a simulation.

\begin{figure*}[t]
\centering
    \includegraphics[width=0.8\textwidth]{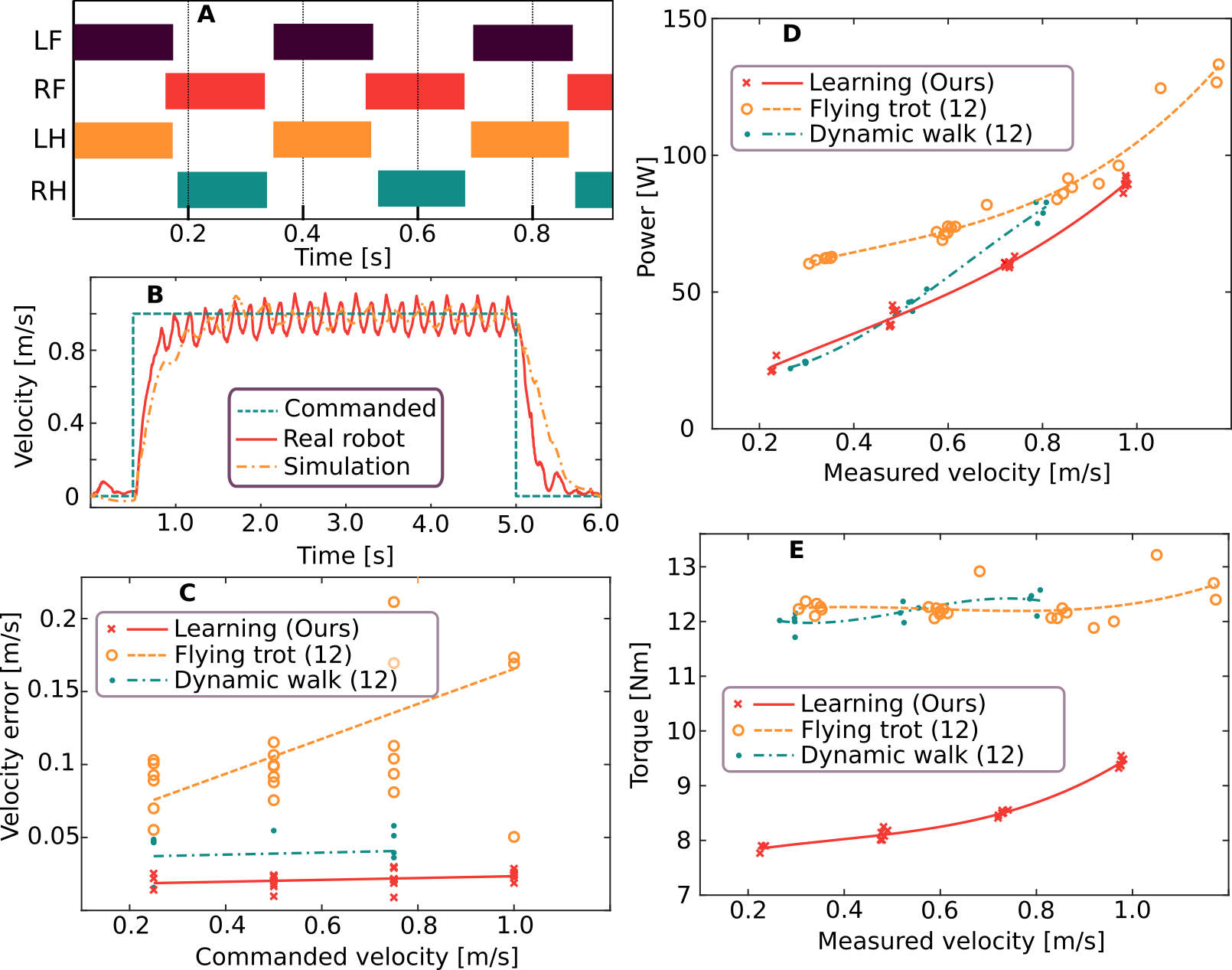}
    \caption{\textbf{Quantitative evaluation of the learned locomotion controller.} (\textbf{A}) The discovered gait pattern for 1.0 m/s forward velocity command. \revise{The abbreviations stand for Left Front (LF) leg, Right Front (RF) leg, Left Hind (LH) leg, and Right Hind (RH) leg, respectively.}  (\textbf{B}) The accuracy of the base velocity tracking with our approach. (\textbf{C})-(\textbf{E}) Comparison of the learned controller against the best existing controller, in terms of power efficiency, velocity error, and torque magnitude, given forward velocity commands of 0.25, 0.5, 0.75, and 1.0 m/s.}
\end{figure*}

Figure~2C, 2D, and 2E compare the performance of the learned controller to the approach of Bellicoso et al.~\cite{bellicoso2018dynamic} in terms of accuracy and efficiency. We used two gaits from~\cite{bellicoso2018dynamic} for the comparison: flying trot, the only gait that can achieve \unit[1]{m/s}, and dynamic lateral walk, the most efficient gait. First, we compare the velocity error at various commanded velocities in Fig.~2C. The learned controller is more accurate than the prior controller for all commanded velocities: by a factor of 1.5 to 2.5 compared to the dynamic lateral walk and by a factor of 5 to 7 compared to the flying trot, depending on the speed. Figure.~2D shows the mechanical power output as a function of the measured velocity. The learned controller performs similarly to the dynamic lateral walk and more efficiently than the flying trot by a factor of 1.2 to 2.5, depending on the speed. Finally, Fig.~2E plots the average measured torque magnitude against the measured velocity. The learned controller is significantly more efficient in this respect than both prior gaits, using \unit[23]{\%} to \unit[36]{\%} less torque depending on the velocity. This large improvement in efficiency is possible since the learned controller walks with 10 to \unit[15]{degree} straighter nominal knee posture. The nominal posture cannot be adjusted to this level in the approach of Bellicoso et al. since this would drastically increase the rate of failure (falling).

Next, we compare our method to ablated alternatives: training with an ideal actuator model and training with an analytical actuator model. The ideal actuator model assumes that all controllers and hardware inside the actuator have infinite bandwidth and zero latency. The analytical model uses the actual controller code running on the actuator in conjunction with identified dynamics parameters from experiments and Computer-Aided Design~(CAD) tools. Some parameters, such as latency, damping, and friction are hand-tuned to increase the accuracy of predicted torque in relation to data obtained from experiments. The policy training procedure for each method is identical to ours.

Both alternative methods could not make a single step without falling. The resulting motions are shown in movies \href{https://youtu.be/WbRXZKUR5Ew}{S4} and \href{https://youtu.be/NYMEA2PD9rQ}{S5}. We observed violent shaking of the limbs, probably due to not accounting for various delays properly. Even though the analytical model contains multiple delay sources that are tuned using real data, accurately modeling all delay sources is complicated when the actuator has limited bandwidth. SEA mechanisms generate amplitude-dependent mechanical response time, and manual tuning of latency parameters becomes challenging. We have tuned the analytical model for more than a week without much success.

\subsection*{High-speed locomotion}
In the previous section, we evaluated the generality and robustness of the learned controller.
Now we focus on operating close to the limits of the hardware to reach the highest possible speed.
The notion of high speed is in general hardware-dependent. There are some legged robots that are exceptional in this regard. Park et al.~\cite{park2015variable} demonstrated full 3D legged locomotion at over \unit[5.0]{m/s} with the MIT Cheetah. The Boston Dynamics WildCat has been reported to reach \unit[8.5]{m/s}~\cite{BDWildCatyoutube}. These robots are designed to run as fast as possible whereas ANYmal is designed to be robust, reliable, and versatile. The current speed record on ANYmal is \unit[1.2]{m/s} and has been set using the flying trot gait~\cite{bellicoso2018dynamic}. Although this may not seem high, it is \unit[50]{\%} faster than the previous speed record on the platform~\cite{gehring2016practice}. Such velocities are challenging to reach via conventional controller design while respecting all limits of the hardware.

We have used the presented methodology to train a high-speed locomotion controller. This controller was tested on the physical system by slowly increasing the commanded velocity to \unit[1.6]{m/s} and lowering it to zero after 10 meters. The forward speed and joint velocities/torques are shown in Fig.~3. ANYmal reached \unit[1.58]{m/s} in simulation and \unit[1.5]{m/s} on the physical system when the command was set to \unit[1.6]{m/s}. All speed values were computed by averaging over at least 3 gait cycles. The controller used both the maximum torque~(\unit[40]{Nm}) and the maximum joint velocities~(\unit[12]{rad/s}) on the physical system as shown in Fig.~3B and 3C. This shows that the learned policy can exploit the full capacity of the hardware to achieve the goal. For most existing methods, planning while accounting for the limitations of the hardware is very challenging, and executing the plan on the real system reliably is harder still. Even state-of-the-art methods \cite{bellicoso2018dynamic,winkler2017fast} cannot limit the actuation during planning due to limitations of their planning module. Modules in their controllers are not aware of the constraints in the later stages and, consequently, their outputs may not be realizable on the physical system.

\begin{figure}
\centering
    \includegraphics[width=0.45\textwidth]{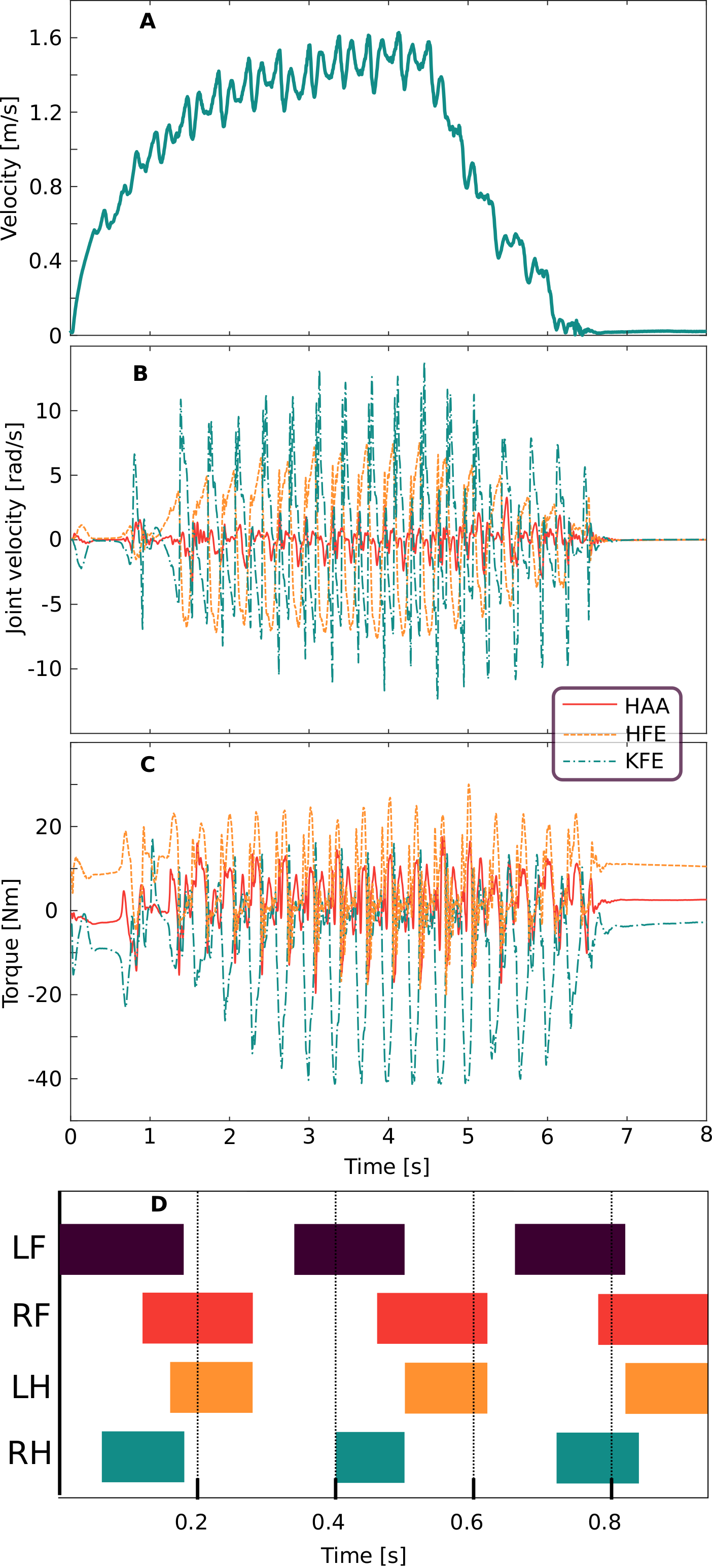}
    \caption{\textbf{Evaluation of the trained policy for high-speed locomotion.} (\textbf{A}) Forward velocity of ANYmal. (\textbf{B}) Joint velocities. (\textbf{C}) Joint torques. (\textbf{D}) Gait pattern.  \revise{The abbreviations stand for Left Front (LF) leg, Right Front (RF) leg, Left Hind (LH) leg, and Right Hind (RH) leg, respectively.}}
\end{figure}

The gait pattern produced by our learned high-speed controller, shown in Fig.~3D, is distinct from the one exhibited by the command-conditioned locomotion controller. It is close to a flying trot but with significantly longer flight phase and asymmetric flight phase duration. This is not a commonly observed gait pattern in nature and we suspect that it is among multiple near-optimal solution modes for this task. The behavior of the policy is illustrated in \href{https://youtu.be/wR3xnK0ZCNs}{movie S6}.

\subsection*{Recovery from a fall}

Legged systems change contact points as they move and are thus prone to falling. If a legged robot falls and cannot autonomously restore itself to an upright configuration, a human operator must intervene. Autonomous recovery after a fall is thus highly desirable. One possibility is to represent recovery behaviors as well-tuned joint trajectories that can simply be replayed: an approach that has been taken in some commercial systems~\cite{shamsuddin2011humanoid}. Such trajectories have required laborious manual tuning. They also take a very long time to execute since they do not take dynamics into account in the motion plan or the control. Some robots are designed such that recovery is either unnecessary or trivial~\cite{saranli2001rhex,ackerman2012boston}. However, such a design may not be possible for bigger and more complex machines. Morimoto et al.~\cite{morimoto2001acquisition} demonstrated that a standing-up motion can be learned on a real robot. However, a simple three-link chain was used for demonstration and the method has not been scaled to realistic systems.

Fast and flexible recovery after a fall, as seen in animals, requires dynamic motion with multiple unspecified contact points.
The collision model for our quadruped is highly complicated: it consists of 41 collision bodies, such as boxes, cylinders, and spheres (Fig.~1, step 1). Planning a feasible trajectory for such a model is extremely complicated. Even simulating such a system is challenging since there are many internal contacts. We use the approach of Hwangbo et al.~\cite{hwangbo2018per} due to its ability to handle such simulation in numerically stable fashion.

Using the presented methodology, we trained a recovery policy and tested it on the real robot. 
We place ANYmal in nine random configurations and activate the controller as shown in \href{https://youtu.be/bbp2vcNb7jg}{movie S7}. Many challenging configurations are tested, including a nearly entirely upside-down configuration (pose 8) and more complex contact scenarios where ANYmal is resting on its own legs (pose 2 and 4).
In all tests, ANYmal successfully flipped itself upright. An example motion is shown in Fig.~4.
These agile and dynamic behaviors demonstrate that our approach is able to learn performant controllers for tasks that are difficult or impossible to address with prior methods.

\begin{figure*}[t]
    \includegraphics[width=1\textwidth]{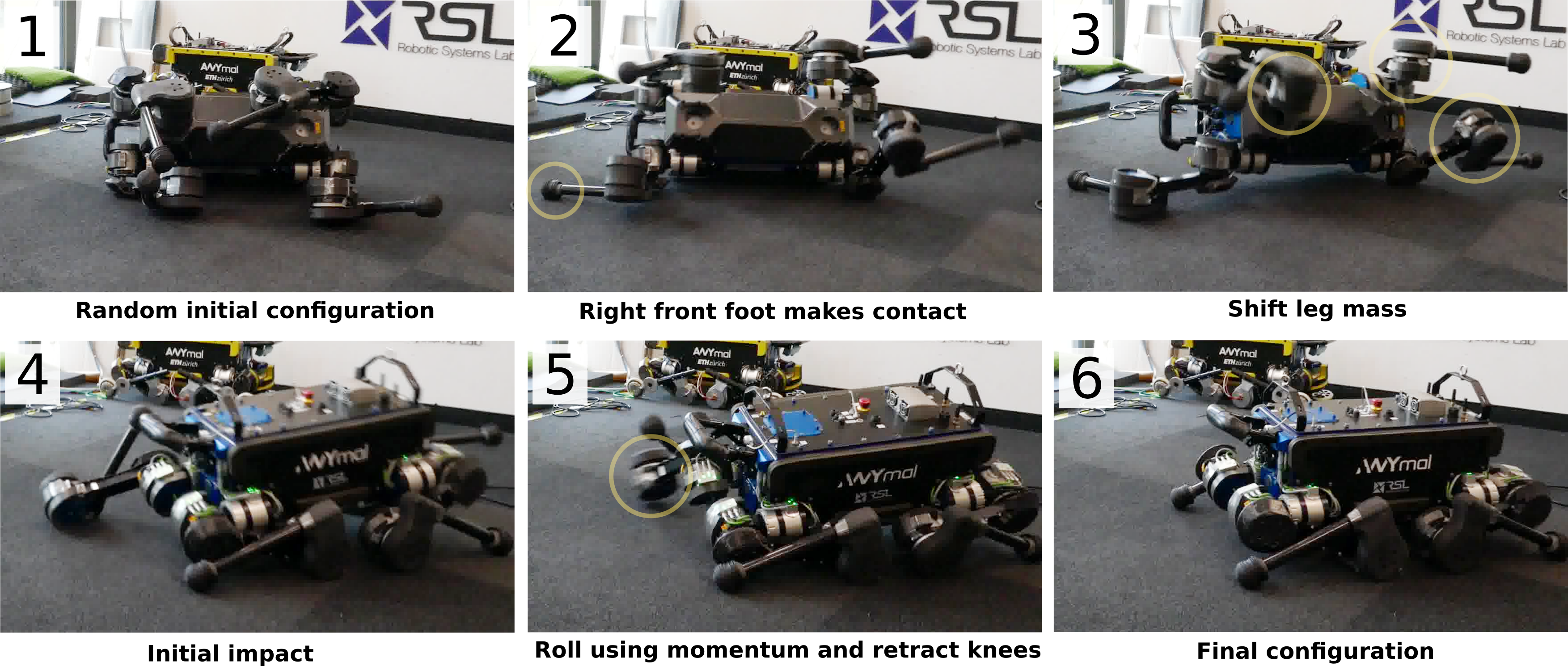}
    \caption{\textbf{A learned recovery controller deployed on the real robot.} The learned policy successfully recovers from a random initial configuration in less than 3 seconds.}
\end{figure*}

\section*{Discussion}
The learning-based control approach presented in this paper achieved a new level of locomotion skill based purely on training in simulation and without tedious tuning on the physical robot. The system achieved more precise and energy-efficient motions than the prior state of the art. It outperformed the previous speed record by \unit[25]{\%} and learned to consistently restore the robot to an operational configuration by dynamically rolling over its body.

Existing controllers are created by engineers. A model with adequate complexity has to be designed and a control strategy has to be developed, tested, and tuned. This process typically takes months and has to be repeated for every distinct maneuver. In contrast, the simulation and learning framework used in this work are applicable to any rigid body system. For applications to new tasks, our method only requires a task description, which consists of the cost function, the initial state distribution, and randomization.


In our method, learned actuator dynamics significantly reduce the reality gap, while stochastic modeling guides the policy to be sufficiently conservative. The recovery task was successful on the very first attempt on the hardware. We then further improved the success rate to \unit[100]{\%} by relaxing the joint velocity constraints. The results presented here were obtained on the second day of experiments on the physical system.
In contrast, due to many model-abstraction layers which are necessary to make the computation tractable, prior methods often cannot exploit a sophisticated actuator model in controlling a complex legged system. Consequently, they often compromise performance or rely on well-tuned low-level controllers. For example, low-level controllers (e.g., the tracking controllers and the whole-body controller) have to be extensively tuned in the tested model-based controller~\cite{bellicoso2018dynamic} to mitigate imperfections of the actuators. 

The learned policies are also robust to changes in hardware, such as those caused by wear and tear. All control policies have been tested for more than three months on the real robot without any modification. Within this period, the robot was heavily used with many controllers, including the ones presented here. Many hardware changes were introduced as well: different robot configurations, which roughly contribute \unit[2.0]{kg} to the total weight, and a new drive which has a spring three times stiffer than the original one. All of the policies presented in this paper have performed robustly even under such conditions.

In terms of computational cost, our approach has an advantage over prior methods. Although it requires several hours of training with an ordinary desktop PC, the inference on the robot requires less than \unit[25]{$\mu s$} using a single CPU thread. Our method shifts nearly all computational costs to the training phase, where we can use external computational resources. Prior controllers often require two orders of magnitude more onboard computation. These demanding requirements limit the level of sophistication and thus the overall performance of the controller.

Using a policy network that directly outputs a joint-level command brings another advantage to our method. In contrast to many prior methods that have numerical issues at singular configurations of the robot, our policies can be evaluated at any configuration. Consequently, our method is free from using ad hoc methods (e.g., branching conditions) in resolving such issues.


While our approach allows for largely automated discovery of performant policies, it still requires some human expertise. A cost function and an initial state distribution have to be designed and tuned for each task. For a person with good understanding on both the task and RL, this process takes about two days for the locomotion policies presented in this work. Although this is still significant amount of time, all the necessary tuning happens in simulation. Therefore, the development time will keep decreasing as computational technology evolves. In contrast, the prior controllers that employ model abstractions inevitably require more development time and often extensive tuning on the real systems.
Developing the recovery policy took about a week largely due to the fact that some safety concerns (i.e., high impacts, fast swing legs, collisions with fragile components, etc) are not very intuitive to embed in a cost function.
Achieving a stand-up behavior was as simple as other tasks. However, for achieving the safe and robust behaviors that are demonstrated in this work, the cost function had to be tweaked several times. Longer development time was also attributed to the fact that it was trained by a person who had no previous experience with any real robot.

To train policies for a new robot, necessary modeling effort has to be made. This includes rigid body modeling using the CAD model and actuator modeling using an actuator network. The former is often automated by modern CAD software and the latter is easy if all necessary software/hardware infrastructures (e.g., logging, regression, and torque measurements) are in place. If not, it also takes a significant portion of the development time. In addition, there are a few actuation types that manifest coupled dynamics (e.g., hydraulic actuators sharing a single accumulator). Learning actuators independently might not result in a sufficient accuracy for these systems. With good understanding on the actuator dynamics, appropriate history configuration can be estimated a priori and tuned further with respect to the validation error. In contrast, constructing an analytical actuator model for ANYmal takes at least three weeks even if there is a very similar model studied in literature~\cite{gehring2016practice}. The model also has many more parameters, many of which cannot be accurately obtained from measurements or the data sheet. Consequently, it requires more tuning than constructing an actuator network.

Another limitation of our approach was observed over the course of this study. A single neural network trained in one session manifests single-faceted behaviors that do not generalize across multiple tasks. Introducing hierarchical structure in the policy network can remedy this and is a promising avenue for future work~\cite{peng2017deeploco}.


\section*{Method}\label{sec:method}
This section describes in detail the simulation environment, the training process, and the deployment on the physical system. An overview of our training method is shown in Fig.~5. The training loop proceeds as follows. The rigid-body simulator outputs the next state of the robot given the joint torques and the current state. The joint velocity and the position error are buffered in a joint state history within a finite time window. The control policy, implemented by a multi-layer perceptron with two hidden layers, maps the observation of the current state and the joint state history to the joint position targets. Finally, the actuator network maps the joint state history and the joint position targets to 12 joint torque values, and the loop continues. In what follows we describe each component in detail.

\subsection*{Modeling rigid-body dynamics}

\begin{figure}[t]
\centering
    \includegraphics[width=0.49\textwidth]{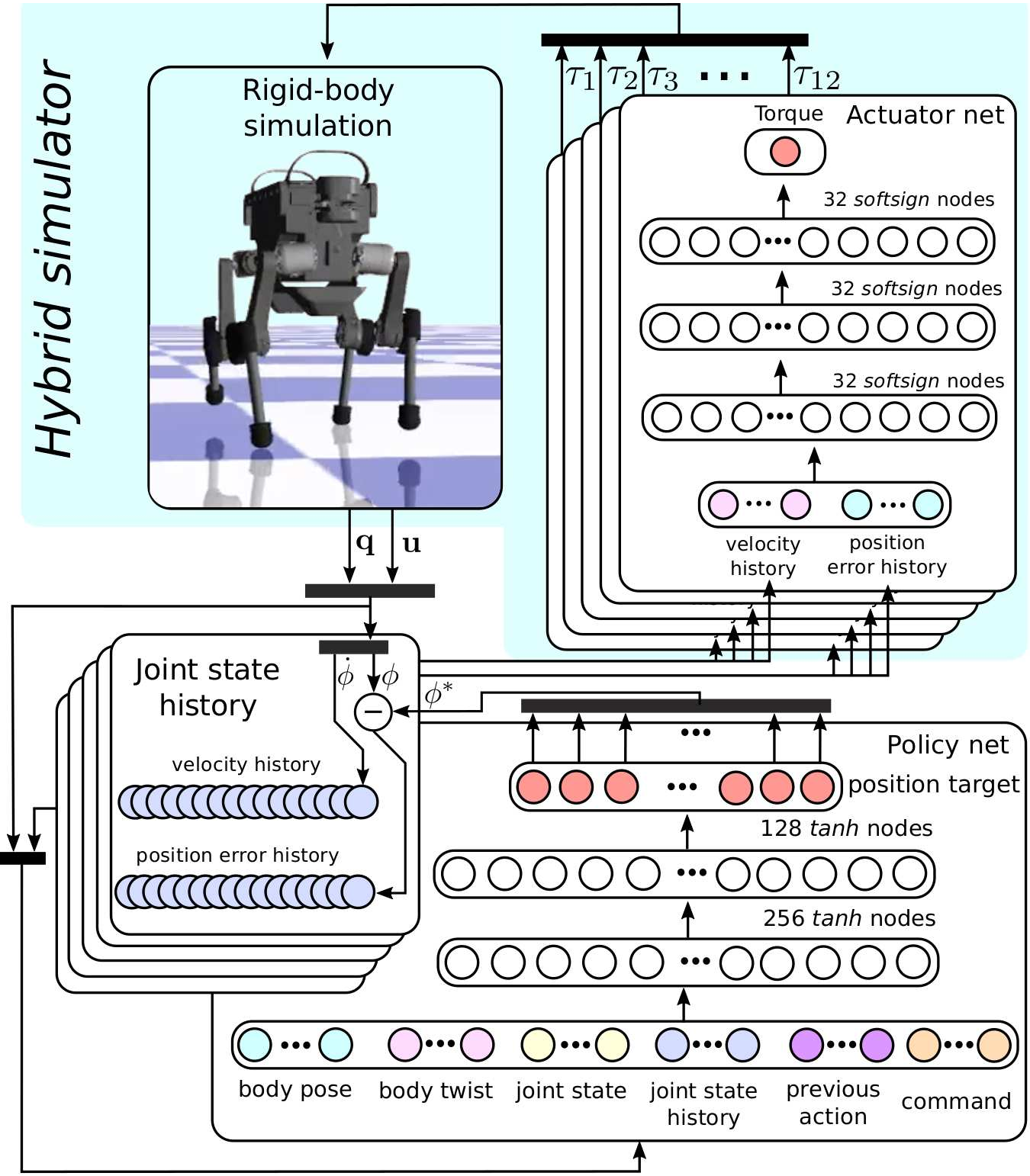}
    \caption{\textbf{Training control policies in simulation.} The policy network maps the current observation and the joint state history to the joint position targets. The actuator network maps the joint state history to the joint torque, which is used in rigid-body simulation. The state of the robot consists of the generalized coordinate $\textbf{q}$ and the generalized velocity $\textbf{u}$. The state of a joint consists of the joint velocity $\dot{\phi}$ and the joint position error, which is the current position $\phi$ subtracted from the joint position target $\phi^*$.}
\end{figure}

To efficiently train a complex policy within a reasonable time and transfer it to the real world, we need a simulation platform that is both fast and accurate. One of the biggest challenges with walking robots is the dynamics at intermittent contacts. To this end, we utilize the rigid body contact solver presented in our previous work~\cite{hwangbo2018per}. This contact solver employs a hard contact model that fully respects the Coulomb friction cone constraint. This modeling technique can accurately capture the true dynamics of a set of rigid bodies making hard contacts with their environment. The solver is not only accurate but also fast, generating about 900,000 time steps per second for the simulated quadruped on an ordinary desktop machine. Because we need hundreds of millions of samples to train a complicated policy, this solver was key to our work.

The inertial properties of the links are estimated from the CAD model. We expect up to about \unit[20]{\%} error in the estimation due to unmodeled cabling and electronics. \revise{To account for such modeling inaccuracies,} we robustify the policy by training with 30 different ANYmal models with stochastically sampled inertial properties. \revise{The center of mass positions, the masses of links, and joint positions are randomized by adding a noise sampled from \unit[$U(-2, 2)$]{cm}, \unit[$U(-15, 15)$]{\%}, and \unit[$U(-2, 2)$]{cm}, respectively.}

\subsection*{Modeling the actuation}
Actuators are an essential part of legged systems. Fast, powerful, lightweight, and high-accuracy actuators typically translate to dynamic, versatile, and agile robots. Most legged systems are driven by hydraulic actuators~\cite{semini2011design} or electric motors with gears~\cite{seok2013design}, and some even include dedicated mechanical compliance~\cite{hutter2016anymal,tsagarakis2013compliant}. These actuators have one thing in common: they are extremely difficult to model accurately. Their dynamics involve nonlinear and nonsmooth dissipation and they contain cascaded feedback loops and a number of internal states that are not even directly observable. Gehring et al.~\cite{gehring2016practice} extensively studied SEA actuator modeling. The model of Gehring et al.\ includes nearly one hundred parameters that have to be estimated from experiments or assumed to be correct from data sheets. This process is error-prone and time-consuming. In addition, many manufacturers do not provide sufficiently detailed descriptions of their products and, consequently, an analytical model may not be feasible.

\revise{To this end, }we use supervised learning to obtain an action-to-torque relationship that includes all software and hardware dynamics within one control loop. More precisely, we train an actuator network that outputs an estimated torque at the joints given a history of position errors (the actual position subtracted from the commanded position) and velocities. \revise{In this work, we assume that the dynamics of the actuators are independent to each other such that we can learn a model for each actuator separately. This assumption might not be valid for other types of actuation. For example, hydraulic actuators with a single common accumulator might manifest coupled dynamics and a single large network, that represents multiple actuators together, might be more desirable.} 

The states of the actuators are only partially observable because the internal states of the actuators \revise{(e.g., states of the internal controllers and motor velocity)} cannot be measured directly. We assume that the network can be trained to estimate the internal states given a history of position errors and velocities, since otherwise the given information is simply insufficient to control the robot adequately. The actuator \revise{used in this work} is revolute and radially symmetric, \revise{and the} absolute angular position is irrelevant given the position error. \revise{We use a history consisting} of the current state and two past states that correspond to $t-0.01$ and $t-0.02$ seconds. \revise{Note that too sparse input configuration might not effectively capture the dynamics at high frequency ($>\unit[100]{Hz}$). This issue is partially mitigated by introducing a smoothness cost term, which penalizes abrupt changes in the output of the policy. Too dense history can also have adverse effects: it is more prone to overfitting and computationally more expensive. The length of the history should be chosen such that it is sufficiently longer than the sum of all communication delays and the mechanical response time. In practice, the exact input configuration is tuned with respect to the validation error. This tuning process often takes less than a day since the network is very small.}

To train the network, we collected a dataset consisting of joint position errors, joint velocities, and the torque. We used a simple parameterized controller that generates foot trajectories in the form of a sine wave; the corresponding joint positions were computed using inverse kinematics. The feet constantly made or break a contact with the ground during data collection so that the resulting trajectories roughly mimicked the trajectories followed by a locomotion controller. To obtain a rich set of data, we varied the amplitude (5$\sim$\unit[10]{cm}) and the frequency (1$\sim$\unit[25]{Hz}) of the foot trajectories and disturbed the robot manually during data collection. We found that the excitation must cover a wide range of frequency spectra since, otherwise, the trained model generates unnatural oscillation even during the training phase. Data collection took less than \unit[4]{min} since the data can be collected in parallel from the 12 identical actuators on ANYmal. Data was collected at 400 Hz, therefore the resulting dataset contains more than a million samples. Approximately \revise{\unit[90]{\%} of the generated data was used for training, and the rest was used for validation.}

The actuator network is a multi-layer perceptron (MLP) with $3$ hidden layers of $32$ units each (Fig.~5, ``Actuator net'' box).
After testing with \revise{two} common \revise{smooth and bounded} activation functions~-- tanh and softsign~\cite{bergstra2009quadratic}~-- we chose the softsign activation function, since it is computationally efficient and provides a smooth mapping. Evaluating the actuator network for all 12 joints takes \unit[12.2]{$\mu s$} with softsign, and \unit[31.6]{$\mu s$} with tanh. As shown here, the tanh activation function results in \revise{a} higher computational cost and is therefore less preferred. \revise{The two} activation functions resulted in approximately the same validation error (0.7$\sim$\unit[0.8]{Nm} in RMS). The validation result with the softsign function is shown in Fig.~6. The trained network nearly perfectly predicts the torque from \revise{the validation data}, whereas the ideal actuator model fails to produce a reasonable prediction. Here the ideal actuator model assumes that there is no communication delay and that the actuator can generate any commanded torque instantly (i.e., infinite actuator bandwidth). The trained model has an average error of \unit[$0.740$]{Nm} \revise{on the validation set}, which is not far from the resolution of the torque measurement~(\unit[$0.2$]{Nm}) and much smaller than the error of the ideal actuator model (\unit[$3.55$]{Nm}). \revise{Its prediction error on test data (i.e., collected using the trained locomotion policies) is significantly higher (\unit[$0.966$]{Nm}) but still far less than that of the ideal model (\unit[$5.74$]{Nm}).}

\begin{figure*}[t]
    \centering
    \includegraphics[width=0.95\textwidth]{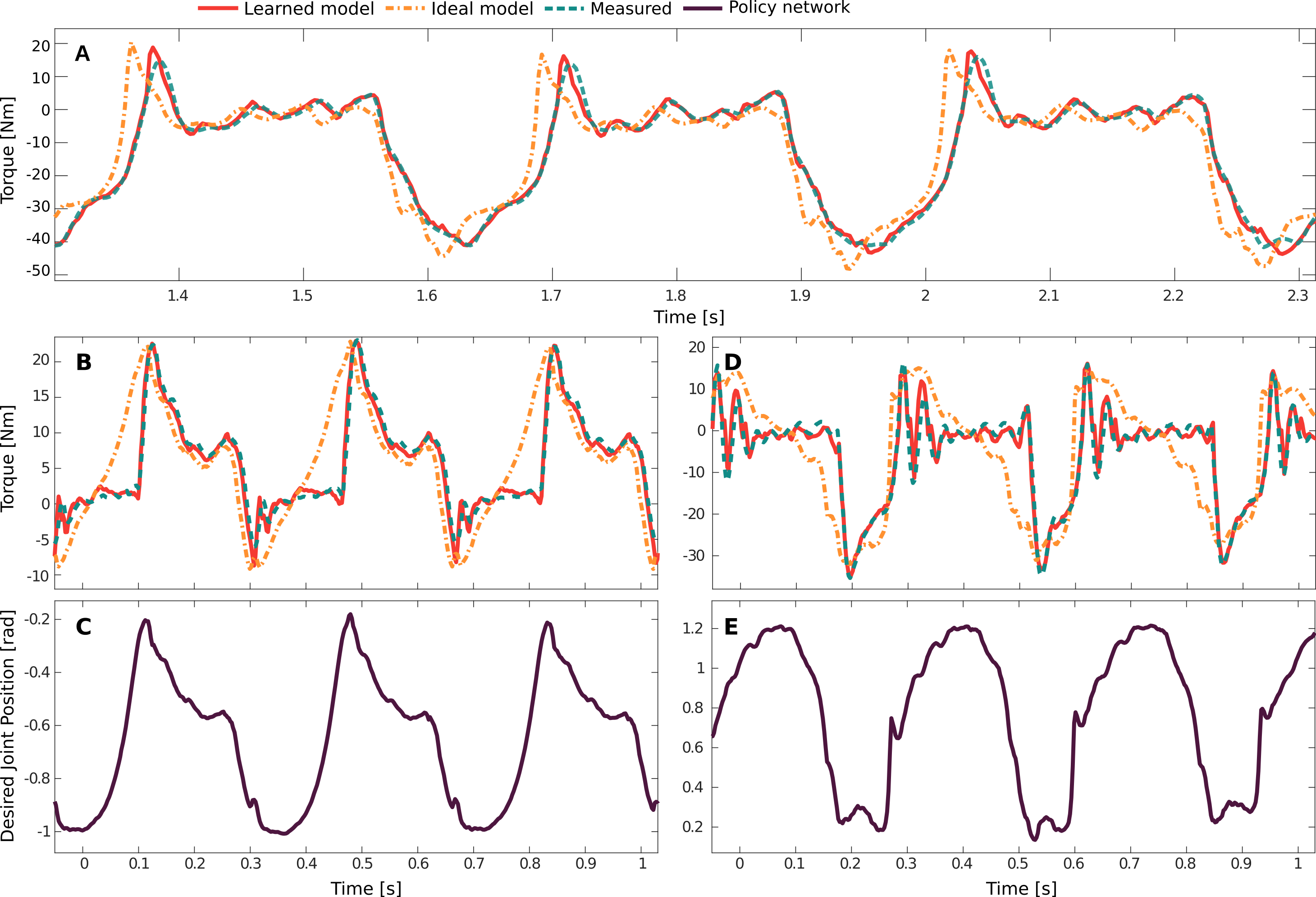}
    \caption{\textbf{Validation of the learned actuator model.} The measured torque and the predicted torque from the trained actuator model are shown. \revise{The ``ideal model'' curve is computed assuming an ideal actuator (i.e., zero communication delay and zero mechanical response time) and is shown for comparison. (\textbf{A}) Validation set (\textbf{B},\textbf{C}) Data from a command-conditioned policy experiment with \unit[0.75]{m/s} forward command velocity and its corresponding policy network output, respectively (\textbf{D},\textbf{E}) Data from a high-speed locomotion policy experiment with \unit[1.6]{m/s} forward command velocity and its corresponding policy network output, respectively.} Note that the measured ground truth in (\textbf{A}) is nearly hidden since the predicted torque from the trained actuator network accurately matches the ground-truth measurements. \revise{Test data is collected at one of the knee joints.}}
\end{figure*}


\subsection*{Reinforcement learning}
We represent the control problem in discretized time.
At every time step $t$ the agent obtains an observation $o_t\in \mathcal{O}$, performs an action $a_t\in \mathcal{A}$, and achieves a scalar reward $r_t\in \mathcal{R}$. We refer to reward and cost interchangeably, with cost being the negative of the reward. We denote by $\mathbf{O}_t = \langle o_t,\, o_{t-1},\, \ldots,\, o_{t-h} \rangle$ the tuple of recent observations. The agent selects actions according to a stochastic policy $\pi(a_t | \mathbf{O}_t)$, which is a distribution over actions conditioned on the recent observations. The aim is to find a policy that maximizes the discounted sum of rewards over an infinite horizon:
\begin{equation}\label{eq:RL}
\pi^* = \argmax_\pi\mathbb{E}_{\boldsymbol{\tau}(\pi)}\left[\sum^{\infty}_{t=0} \gamma^t r_t\right],
\end{equation}
where 
$\gamma \in (0,1)$ is the discount factor and $\vect{\tau}(\pi)$ is the trajectory distribution under policy $\pi$ (the distribution depends both on the policy and the environment dynamics). 
In our setting, the observations are the measurements of robot states provided to the controller, the actions are the position commands to the actuators, and the rewards are specified so as to induce the behavior of interest.


A variety of reinforcement learning algorithms can be applied to the specified policy optimization problem. We chose Trust Region Policy Optimization (TRPO)~\cite{schulman2015trust}, a policy gradient algorithm that has been demonstrated to learn locomotion policies in simulation~\cite{schulman2016high}. It requires almost no parameter tuning; we use only the default parameters (as provided in~\cite{schulman2015trust, schulman2016high}) for all learning sessions presented in this paper. We used a fast custom implementation of the algorithm~\cite{hwangbo2017control}. This efficient implementation and fast rigid-body simulation~\cite{hwangbo2018per} allowed us to generate and process about a quarter of a billion state transitions in roughly four hours. A learning session terminates if the average performance of the policy does not improve by more than a task-specific threshold within 300 TRPO iterations.

\subsection*{Observation and action}
The observations in our method should be observable (i.e., can be inferred from measurements) on the real robot and relevant for the task. The joint angles, velocities, and body twist are all observable and highly relevant. Measuring the body orientation is not straightforward since only two degrees of freedom in the orientation are observable with an Inertial Measurement Unit~(IMU). The set of observable degrees in the orientation is in bijection with $S^2$, or with a unit vector, which can be interpreted as the direction of the gravity vector expressed in the IMU frame. We denote this unit vector as $\vect{\phi}^g$. The height of the base is not observable, but we can estimate it from the leg kinematics, assuming the terrain is flat. A simple height estimator based on a 1D Kalman filter was implemented along with the existing state estimation \cite{bloesch2013state}. However, since this height estimator cannot be used when the robot is not on its feet, we removed the height observation when training for recovery from a fall. The whole observation at $t=t_k$ is defined as $o_k=\langle\vect{\phi}^g, r_z, v,\omega, \phi, \dot{\phi}, \vect{\Theta}, a_{k-1}, C \rangle$, where $r_z$, $v$, and $\omega$ are height, linear, and angular velocities of the base, $\phi$ and $\dot{\phi}$ are positions and velocities of the joints, $\vect{\Theta}$ is a sparsely sampled joint state history, $a_{k-1}$ is the previous action, and $C$ is the command. The joint state history is sampled at $t=t_k-\unit[0.01]{s}$ and $t=t_k-\unit[0.02]{s}$.

The joint state history was essential in training a locomotion policy. We hypothesize that this is due to the fact that it enables contact detection. An alternative way to detect contacts is to use force sensors which give a reliable contact state estimate. However, such sensors increase the weight of the end-effectors and consequently lower the energy efficiency of the robot. \revise{The exact history configuration was found empirically by analyzing the final performance of the policy.}

\revise{Our policy outputs low-impedance joint position commands, which we find to be very effective in many tasks.} Peng et al.~\cite{peng2017learning} found that \revise{such a controller} can outperform a torque controller in both training speed and final control performance. Even though there is always a bijective map between them, the two action parameterizations have different smoothness and thus different training difficulty. In addition, a position policy has an advantage in training since it starts as a standing controller whereas a torque controller initially creates many trajectories that result in falling. Thus \revise{we use the policy network as an impedance controller. Our network outputs a single position reference, which is converted to torque using fixed gains ($k_p=\unit[50]{Nm/rad}$ and $k_d=\unit[0.1]{Nm/rad/s}$) and zero target velocity.} \revise{The position gain is chosen to be roughly the nominal range of torque ($\pm$\unit[30]{Nm}) divided by the nominal range of motion ($\pm$\unit[0.6]{rad}). This ensures that the policy network has similar output range for torque and position. The velocity gain is chosen to be sufficiently high to prevent unwanted oscillation on the real robot. From our experience, the final locomotion performance is robust against a small change in gains. For instance, increasing the position gain to \unit[80]{Nm/rad} does not noticeably change the performance.}

Note that the position policy we use here is different from position controllers commonly used in robotics. Position controllers are sometimes limited in performance when the position reference is time-indexed, which means that there is a higher-level controller that assumes that the position plan will be followed at high accuracy. This is the main reason that torque controllers have become popular in legged robotics. However\revise{, as in many other RL literature,} our control policy is state-indexed and does not suffer from the limitations of common PD controllers. The policy is trained to foresee that position errors will occur and even uses them to generate acceleration and interaction forces. \revise{In addition, thanks to kinematic randomization, a trained policy does not solely rely on kinematics: the policy inevitably has to learn to exert appropriate impulse on the environment for locomotion. This makes our policy more robust since impulse-based control approaches are known to be more robust against system changes and model inaccuracies~\cite{park2015variable}.}

\subsection*{Policy training details} \label{sec:method_training}
The control policies presented in this work were trained only in simulation. \revise{In order to train performant policies using only simulated data, we follow both standard and problem-specific training procedures. Here we describe them in detail and explain the rationale behind them.}

\revise{Training control policies for locomotion have been demonstrated multiple times in literature.~\cite{schulman2015trust,heess2017emergence,peng2017deeploco}. However, many of the trained policies do not manifest natural motions and it is highly questionable if they will work on physical systems. Some researchers have noticed that naive methods cannot generate natural-looking and energy-efficient locomotion behaviors~\cite{yu2018learning}. Low penalty on joint torque and velocity results in unnatural motions whereas high penalty on them results in a standing behavior. The main reason for the standing behavior is that such a behavior is already a good local minimum when there is high penalty associated with motion.}

We solved this problem by introducing a curriculum: \revise{using a curriculum, we shape the initial cost landscape such that the policy is strongly attracted to a locomotion policy and then later polish the motion to satisfy the other criteria.} A simple curriculum was generated by modulating the coefficients of the cost terms and the disturbance via a multiplicative curriculum factor. \revise{We define a} curriculum factor {which} describes the progression of the curriculum: $k_c = k_0\in(0,1)$ corresponds to the start of the curriculum and $k_c = 1$ corresponds to the final difficulty level. The intermediate values are computed as $k_{c,j+1} \leftarrow (k_{c,j})^{k_d}$, 
where $k_d\in(0,1)$ is the advance rate, which describes how quickly the final difficulty level is reached, and $j$ is the iteration index of RL training. \revise{The sequence of curriculum factors is monotonically increasing and asymptotically converging to 1 within the given parameter intervals. We suspect that many other update rules adhering to these criteria will result in similar learning performance.} All of cost terms are multiplied by this curriculum factor, except the cost terms related to the objective (i.e., base velocity error cost in the command-conditioned and high-speed locomotion task and base orientation cost in recovery task). This way, the robot first learns how to achieve the objective and then how to respect various constraints. \revise{This technique is related to curriculum learning introduced by Bengio et al.~\cite{bengio2009curriculum}, which incrementally introduces samples of more difficulties. Instead of altering the samples, we alter the objective to control the training difficulty.} \revise{For all training sessions, we use $k_0 = 0.3$ and $k_d = 0.997$. The parameter $k_0$ should be chosen to prevent the initial tendency to stand still. It can be easily tuned by observing the first one hundred iterations of the RL algorithm. The parameter $k_d$ is chosen such that the curriculum factor almost reaches 1 (or $\sim 0.9$) at the end of training. Although the required number iterations are not known a priori, there are sufficient publications on RL applications (including this one) to provide necessary insights to the users.} 

\revise{We tuned the discount factor $\gamma$ (Eq.~\eqref{eq:RL}) separately for each task based on the qualitative performance of the trained controllers in simulation. For training the command-conditioned controller and the high-speed controller, we used $\gamma = 0.9988$ which corresponds to a half-life of \unit[5.77]{s}. We also successfully trained almost equally performant policies with lower half-life ($\sim\unit[2]{s}$) but they manifest a less natural standing posture. For training the recovery controller, we used $\gamma = 0.993$, which corresponds to a half-life of \unit[4.93]{s}. A sufficiently high discount factor shows more natural standing posture due to the fact that it penalizes standing torque more than motion (torque, joint velocities and other quantities incurring due to motion). However, too high discount factor might result in a slow convergence so it should be tuned appropriately depending on the task.} For training command-conditioned and high-speed locomotion, TRPO finished training in nine days of simulated time, which corresponds to four hours of computation in real time. For training for recovery from a fall, TRPO took \unit[79]{days} of simulated time, which corresponds to eleven hours of computation in real time.

For command-conditioned and high-speed locomotion, we represent a command by three desired body velocity values: forward velocity, lateral velocity, and the turning rate.
During training, the commands are sampled randomly from predefined intervals (see tables S1 and S2 for details) and the cost defined in section S3 is employed.
The initial state of the robot is sampled either from a previous trajectory or a random distribution, shown in table S3, with equal probability. 
This initialization procedure generates data containing complicated state transitions and robustifies the trained controller.
Each trajectory lasts 6 seconds unless the robot reaches a terminal state earlier. There are two possibilities for termination: violating joint limits and hitting the ground with the base. \revise{Upon termination, agent receives a cost of 1 and is reinitialized. The value of the termination cost is not tuned: since only the ratio between the cost coefficients is important for the final performance, we tune other cost terms to work with this terminal value.}

For training recovery from a fall, the collision bodies \revise{of the ANYmal model} are randomized in size and position. Samples that result in unrealistic internal collisions are removed. The cost function and the initial state distribution are described in section S4 and fig.~S3, respectively. \revise{The special initialization method in section S4} is needed to train for this task since naive sampling \revise{often} results in interpenetration and the dynamics become unrealistic. To this end, we dropped ANYmal from a height of \unit[1.0]{m} with randomized orientations and joint positions, ran the simulation for \unit[1.2]{s}, and used the resulting state as initialization.

Another crucial detail is that joint velocities cannot be directly measured on the real robot. 
Rather, they are computed by numerically differentiating the position signal, which results in noisy estimates.
We model this imperfection by injecting a strong additive noise (\unit[$U(-0.5,0.5)$]{rad/s}) \revise{to the joint velocity measurements} during training.
This way we ensure that the learned policy is robust to inaccurate velocity measurements.
We also add noise during training to the observed linear velocity (\unit[$U(-0.08,0.08)$]{m/s}) and angular velocity (\unit[$U(-0.16,0.16)$]{m/s}) of the base. The rest of the observations are noise-free.
Interestingly, removing velocities from the observation altogether led to a complete failure to train, even though in theory the policy network could infer velocities as finite differences of observed positions. \revise{We explain this by the fact that non-convexity of network training makes appropriate input pre-processing important.} For similar reasons, input normalization is necessary in most learning procedures.

We implemented the policy with an MLP with two hidden layers, with 256 and 128 units each and tanh nonlinearity (Fig.~5). We found that the nonlinearity has a strong effect on performance on the physical system. Performance of two trained policies with different activation functions can be very different in the real world even when they perform similarly in simulation. Our explanation is that unbounded activation functions, such as ReLU, can degrade performance on the real robot, since actions can have very high magnitude when the robot reaches states that were not visited during training.

Bounded activation functions, such as tanh, yield less aggressive trajectories when subjected to disturbances. \revise{We believe this is true for softsign as well, but it is not tested in policy networks due to an implementation issue in our RL framework~\cite{hwangbo2017control}.}

\subsection*{Deployment on the physical system}
We use the ANYmal robot~\cite{hutter2016anymal}, shown in step four of Fig.~1, to demonstrate the real-world applicability of our method. ANYmal is a dog-sized quadrupedal robot weighing about \unit[32]{kg}. Each leg is about \unit[55]{cm} long and has three actuated degrees of freedom, namely Hip Abduction/Adduction (HAA), Hip Flexion/Extension (HFE), and Knee Flexion/Extension (KFE).

ANYmal is equipped with 12 SEAs~\cite{pratt1995series, Hutter2016patentAnydrive}. An SEA is composed of an electric motor, a high gear ratio transmission, an elastic element, and two rotary encoders to measure spring deflection and output position. In this work, we use a joint-level PD controller with low feedback gains on the joint-level actuator module of the ANYmal robot. The dynamics of the actuators contain multiple components in succession, as follows. First, the position command is converted to the desired torque using a PD controller. Subsequently, the desired current is computed using a PID controller from the desired torque. The desired current is then converted to phase voltage using a Field-Oriented Controller~(FOC), which produces the torque at the input of the transmission. The output of the transmission is connected to an elastic element whose deflection finally generates torque at the joint~\cite{gehring2016practice}. These highly complex dynamics introduce many hidden internal states that we do not have direct access to and complicate our control problem.

After acquiring a parameter set for a trained policy from our hybrid simulation, the deployment on the real system was straightforward. A custom MLP implementation and the trained parameter set were ported to the robot's onboard PC. This network was evaluated at \unit[200]{Hz} for command-conditioned/high-speed locomotion and at 100 Hz for recovery from a fall. We found that performance was surprisingly insensitive to the control rate. For example, the recovery motion was trained at \unit[20]{Hz} but performance was identical when we increased the control rate up to \unit[100]{Hz}.  \revise{This was possible since the flip-up behaviors involve low joint velocities (mostly below \unit[6]{rad/s}). More dynamic behaviors (e.g., locomotion) often require a much higher control rate in order to have an adequate performance. Higher frequency (\unit[100]{Hz})} was used for experiments because it made less audible noise. Even at \unit[100]{Hz}, evaluation of the network uses only \unit[0.25]{\%} of the computation available on a single CPU core.

\section*{Conclusion}
Controllers presented in this paper, trained in a few hours in simulation, outperformed the best existing model-based \revise{controller running on the same robot}, which were designed and tuned over many years. Our learned locomotion policies ran faster and with higher precision while using less energy, torque, and computation. The recovery controller exhibits dynamic roll-over involving multiple \revise{unspecified} contacts with the environment; such a behavior has not been achieved \revise{on a real robot of comparable complexity} with any of the existing optimization schemes. 

\revise{The presented approach is not fundamentally limited to known and simple environments.}
We see the results presented in this paper as a step towards comprehensive locomotion controllers for resilient and versatile legged robots.



\section*{Supplementary materials}
Section S1. Nomenclature \\
Section S2. Random command sampling method employed for evaluating the learned command-conditioned controller.\\
Section S3. Cost terms for training command-conditioned locomotion and high-speed locomotion tasks \\
Section S4. Cost terms for training recovery from a fall \\
Table S1. Command distribution for training command-conditioned locomotion \\
Table S2. Command distribution for training high-speed locomotion \\
Table S3. Initial state distribution for training both the command-conditioned and high-speed locomotion \\
Fig.~S1. Base velocity tracking performance of the learned controller while following random commands from a joystick. \\
Fig.~S2. Base velocity tracking performance of the best existing method while following random commands from a joystick. \\
Fig.~S3. Sampled initial states for training recovery controller \\
Movie S1. Summary of the results and the method.\\
Movie S2. Locomotion policy trained with a learned actuator model.\\
Movie S3. Random command experiment.\\
Movie S4. Locomotion policy trained with an analytical actuator model.\\ 
Movie S5. Locomotion policy trained with an ideal actuator model.\\
Movie S6. Performance of a learned high-speed policy. \\
Movie S7. Performance of a learned recovery policy. \\

\bibliographystyle{Science}


\section*{Acknowledgments}
\textbf{Acknowledgment} We thank ANYbotics for responsive support on ANYmal \textbf{Funding} The project was funded in part by the Intel Network on Intelligent Systems and the Swiss National Science Foundation (SNF) through the National Centre of Competence in Research Robotics and Project 200021-166232. The work has been conducted as part of ANYmal Research, a community to advance legged robotics. \textbf{Author contribution}
J.H conceived the main idea of the train and control methods, set up the simulation and trained networks for the command-conditioned locomotion and the high-speed locomotion. J.L trained a network for recovery from a fall. J.H, A.D, M.H and V.K refined ideas and contributed in the experiment design. J.H and J.L performed experiments together. D.B and V.T helped setting up the hardware for the experiments. J.H, A.D, M.H and V.K analyzed the data and prepared the manuscript. \textbf{Conflict of interest}
The authors declare that they have no competing interests. \textbf{Data and material availability}
All data needed to evaluate the conclusions in the paper are present in the paper or the Supplementary Materials. Other materials can be found at \url{https://github.com/junja94/anymal_science_robotics_supplementary}.



\clearpage

\section*{Supplementary materials}

\subsection*{S1. Nomenclature}

\newcommand{\tab}[1]{\hspace{.05\textwidth}\rlap{#1}}

\makebox[1.2cm]{$k_c$} curriculum factor.\\
\makebox[1.2cm]{$c_\cdot$} coefficient term for a cost term.\\
\makebox[1.2cm]{${{v}}^C_{AB}$} linear velocity of $B$ respect to $A$ expressed in $C$\\
\makebox[1.2cm]{$\omega$} angular velocity\\
\makebox[1.2cm]{$\hat{\cdot}$} desired quantity\\
\makebox[1.2cm]{$\tau$} joint torque\\
\makebox[1.2cm]{$\phi$} angular quantity\\
\makebox[1.2cm]{$v_f$} linear velocity of a foot\\
\makebox[1.2cm]{$v_{ft}$} tangential velocity of a foot (x, y components)\\
\makebox[1.2cm]{$p_f$} linear position of a foot\\
\makebox[1.2cm]{$v_{c,n}$} linear velocity of the n-th contact point\\
\makebox[1.2cm]{$i_{c,n}$} contact impulse of the n-th contact\\
\makebox[1.2cm]{$g_i$} gap function of the i-th possible contact pair\\
\makebox[1.2cm]{$I_{c}$} index set of all contacts\\
\makebox[1.2cm]{$I_{c,f}$} index set of foot contacts\\
\makebox[1.2cm]{$I_{c,i}$} index set of internal contacts\\
\makebox[1.2cm]{$\lvert \cdot \rvert$} cardinality of a set or $l_1$ norm\\
\makebox[1.2cm]{$\lvert\lvert \cdot \rvert\rvert$} $l_2$ norm\\
\makebox[1.2cm]{$\textbf{0}^n$} n-dimensional vector of zeroes\\
\makebox[1.2cm]{$\textbf{q}$} generalized coordinate\\
\makebox[1.2cm]{$\textbf{u}$} generalized velocity\\

\subsection*{S2. Random command sampling method employed for evaluating the learned command-conditioned controller.}

The motivation of having a special sampling method is the limited size of the experimental area. A sufficiently long unconstrained sequence of randomly sampled velocity commands may drive the robot outside the limits of the physical space available. We therefore use the following sampling scheme. We first sample a command from the distribution described in table S1. Then we simulate a position trajectory by assuming that the body follows the velocity command perfectly. If the position of the body goes outside the perimeter of the available space, we reject the sampled command, reset the position to the previous position, and resample a velocity command. This loop continues until the desired number of commands are sampled.

\subsection*{S3. Cost terms for training command-conditioned locomotion and high-speed locomotion}

We used a \revise{logistic} kernel to define a bounded cost function $K: \mathbb{R} \rightarrow [-0.25, 0)$ as
 \begin{equation}
     K(x) = - \frac{1}{e^x + 2 + e^{-x}}.
\label{kernel}
 \end{equation}
\revise{This kernel converts a tracking error to a bounded reward. We found it more useful than Euclidean norm, which is a more common choice. An Euclidean norm generates a high cost in the beginning of training where the tracking error is high such that termination (i.e. falling) becomes more rewarding strategy. On the other hand, the logistic kernel ensures that the cost is lower-bounded by zero and termination becomes less favorable. Many other bell-shaped kernels~(Gaussian, triweight, biweight, etc) have the same functionality and can be used instead of a logistic kernel.} 

The symbols used in this section are defined in section S1. \revise{Note that many cost terms are also multiplied by the time step $\Delta t$ since we are interested in the integrated value over time. Explanation on the curriculum factor $k_c$ can be found in subsection "Training in simulation".}

\vspace{0.15cm}
\noindent \textbf{angular velocity of the base cost} \revise{($c_w = -6\Delta t$)}
\begin{equation}
 c_{w} K(|{\omega}^I_{IB}-\hat{{\omega}}^I_{IB}|)
\end{equation}
\vspace{0.15cm}
\textbf{linear velocity of the base cost} 
\revise{($c_{v1} = -10 \Delta t, c_{v2} = -4 \Delta t$)}
\begin{equation}
    c_{v1} K(|c_{v2}\cdot({v}^I_{IB}-\hat{{v}}^I_{IB})|)
\end{equation}
\vspace{0.15cm}
\textbf{torque cost} \revise{($c_{\tau} = 0.005 \Delta t$)}
\begin{equation}
    k_c c_{\tau}||\tau||^{2}
\end{equation}
\vspace{0.15cm}
\textbf{joint speed cost} \revise{($c_{js} = 0.03 \Delta t$)}
\begin{equation}
    k_c c_{js} \lvert\lvert \dot{{\phi}}^{i} \rvert\rvert ^2 \quad \forall i \in \{1,2...,12\}
\end{equation}
\vspace{0.15cm}
\textbf{foot clearance cost} \revise{($c_{f} = 0.1 \Delta t, \hat{p}_{f,i,z} = 0.07~\textrm{m}$)}
\begin{equation}
    k_c c_{f}(\hat{p}_{f,i,z} - p_{f,i,z})^2||v_{ft,i}||, ~\forall i, g_i > 0, i\in\{0,1,2,3\},
\end{equation}
\vspace{0.15cm}
\textbf{foot slip cost} \revise{($c_{fv} = 2.0 \Delta t$)}
\begin{equation}
    k_c c_{fv}||v_{ft,i}||, ~\forall i, g_i = 0, i\in\{0,1,2,3\}
\end{equation}
\vspace{0.15cm}
\textbf{orientation cost} \revise{($c_{o} = 0.4 \Delta t$)}
\begin{equation}
    k_c c_{o}||[0,0,-1]^T - \phi_{g}||
\end{equation}
\vspace{0.15cm}
\textbf{smoothness cost} \revise{($c_{s} = 0.5 \Delta t$)}
\begin{equation}
    k_c c_{s}||\tau_{t-1} - \tau_{t}||^2
\end{equation}

\subsection*{S4. Cost terms for training recovery from a fall}

We use $\texttt{angleDiff}: \mathbb{R} \times \mathbb{R} \rightarrow [0, \pi] $ that computes the minimum angle difference between two angular positions to define a cost function on the joint positions.
The symbols used in this section are defined in section S1.
\vspace{0.15cm}
\noindent
\textbf{torque cost} \revise{($c_{\tau} = 0.0005 \Delta t$)}
\begin{equation}
    k_c c_{\tau}||\tau||^2
\end{equation}
\vspace{0.15cm}
\textbf{joint speed cost} \revise{($c_{js} = 0.2 \Delta t$, $c_{jsmax} = 8~\unit{rad/s})$}
\begin{equation}
\text{If } \lvert \dot{{\phi}}^{i} \rvert > \lvert c_{jsmax} \rvert, \quad
    k_c c_{js} \lvert\lvert \dot{{\phi}}^{i} \rvert\rvert ^2 \quad \forall i\in \{1,2...12\}
\end{equation}
\vspace{0.15cm}
\textbf{joint acceleration cost}  \revise{($c_{ja} = 0.0000005 \Delta t$)}
\begin{equation}
    k_c c_{ja} \lvert\lvert \ddot{{\phi}}^{i} \rvert\rvert ^2 \quad \forall i\in \{1,2...12\}
\end{equation}
\vspace{0.15cm}
\textbf{HAA cost} \revise{($c_{HAA} = 6.0 \Delta t$)}
\begin{equation}
\text{If } \lvert {\phi}_{roll} \rvert < 0.25 \pi, \quad
    k_c c_{HAA}K(\texttt{angleDiff}({\phi}^{HAA}, 0 ))
\end{equation}
\vspace{0.15cm}
\textbf{HFE cost}  \revise{($c_{HFE} = 7.0 \Delta t$, $\hat{{\phi}}^{HFE} = \pm 0.5 \pi~\unit{rad}$~($+$ for right legs) )}
\begin{equation}
\text{If } \lvert {\phi}_{roll} \rvert < 0.25 \pi, \quad
    k_c c_{HFE}K(\texttt{angleDiff}({\phi}^{HFE}, \hat{{\phi}}^{HFE}))
\end{equation}
\vspace{0.15cm}
\textbf{KFE cost} \revise{($c_{KFE} = 7.0 \Delta t$, $\hat{{\phi}}^{KFE} = \mp 2.45~\unit{rad}$)}
\begin{equation}
\text{If } \lvert {\phi}_{roll} \rvert < 0.25 \pi, \quad
    k_c c_{KFE}K(\texttt{angleDiff}({\phi}^{KFE}, \hat{{\phi}}^{KFE}))
\end{equation}
\vspace{0.15cm}
\textbf{contact slip cost} \revise{($c_{cv} = 6.0 \Delta t$)}
\begin{equation}
    k_c c_{cv} \frac{\sum_{n \in {I_{c}}}||v^I_{c,n}||^2}{\lvert I_{c} \rvert}
\end{equation}
\vspace{0.15cm}
\textbf{body contact impulse cost} \revise{($c_{cimp} = 6.0 \Delta t$)}
\begin{equation}
    k_c c_{cimp} \frac{\sum_{n \in I_{c} \backslash I_{c,f} }||i^I_{c,n}||}{\lvert I_{c} \rvert - \lvert I_{c,f} \rvert}
\end{equation}
\vspace{0.15cm}
\textbf{internal contact cost} \revise{($c_{cint} = 6.0 \Delta t$)}
\begin{equation}
    k_c c_{cint} \lvert I_{c,i} \rvert
\end{equation}
\vspace{0.15cm}
\textbf{orientation cost} \revise{($c_{o} = 6.0 \Delta t$)}
\begin{equation}
    c_{o}||[0,0,-1]^T - \phi_{g}||^2
\end{equation}
\vspace{0.15cm}
\textbf{smoothness cost}  \revise{($c_{s} = 0.0025 \Delta t$)}
\begin{equation}
    k_c c_{s}||\tau_{t-1} - \tau_{t}||^2
\end{equation}

\clearpage

\makeatletter 
\renewcommand{\thetable}{S\@arabic\c@table}
\makeatother

\begin{table}
\centering
\begin{tabular}{ l | l | l }
  \hline
  & min & max \\\hline
  forward velocity & \unit[-1.0]{m/s} & \unit[1.0]{m/s} \\\hline
  lateral velocity & \unit[-0.4]{m/s} & \unit[0.4]{m/s} \\\hline
  turning rate & \unit[-1.2]{rad/s} & \unit[1.2]{rad/s} \\
  \hline  
\end{tabular}
\caption{\textbf{Command distribution for the command-conditioned locomotion task.} During training the command was varied randomly as shown in this table. The range was selected to match the capabilities of the existing controllers.}
\label{tab:commandgl}
\end{table}

\begin{table}
\centering
\begin{tabular}{ l | l | l }
  \hline
  & min & max \\\hline
  forward velocity & \unit[-1.6]{m/s} & \unit[1.6]{m/s} \\\hline
  lateral velocity & \unit[-0.2]{m/s} & \unit[0.2]{m/s} \\\hline
  turning rate & \unit[-0.3]{rad/s} & 0.\unit[3]{rad/s} \\
  \hline  
\end{tabular} \label{tab:commandhs}
\caption{\textbf{Command distribution for the high-speed locomotion task.} During training the command was varied randomly as shown in this table. Only the forward velocity command has a large variation since this task focuses only on high speed.}
\end{table}

\begin{table}
\centering
\begin{tabular}{ l | l | l }
  \hline
  & mean & standard deviation \\ \hline
  base position & $[0, 0, 0.55]^T$ & \unit[1.5]{cm} \\ \hline
  base orientation & $[1, 0, 0, 0]^T$ & \unit[0.06]{rad} (about a random axis) \\ \hline
  joint position & $[0, 0.4, -0.8, 0, 0.4, -0.8,$ & \unit[0.25]{rad} \\
   & $ 0, -0.4, 0.8, 0, -0.4, 0.8]^T$ &  \\ \hline
  base linear velocity & $\textbf{0}^3$ & \unit[0.012]{m/s} \\ \hline
  base angular velocity & $\textbf{0}^3$ & \unit[0.4]{rad/s} \\ \hline
  joint velocity & $\textbf{0}^{12}$ & \unit[2]{rad/s} \\ \hline
\end{tabular} 
\caption{\textbf{Initial state distribution for training the command-conditioned and high-speed locomotion controllers.} The initial state is randomized to make the trained policy more robust.}
\label{tab:initial}
\end{table}

\clearpage

\setcounter{figure}{0}
\makeatletter 
\renewcommand{\thefigure}{S\@arabic\c@figure}
\makeatother

\begin{sidewaysfigure*}
\centering
    \includegraphics[width=0.8\textwidth]{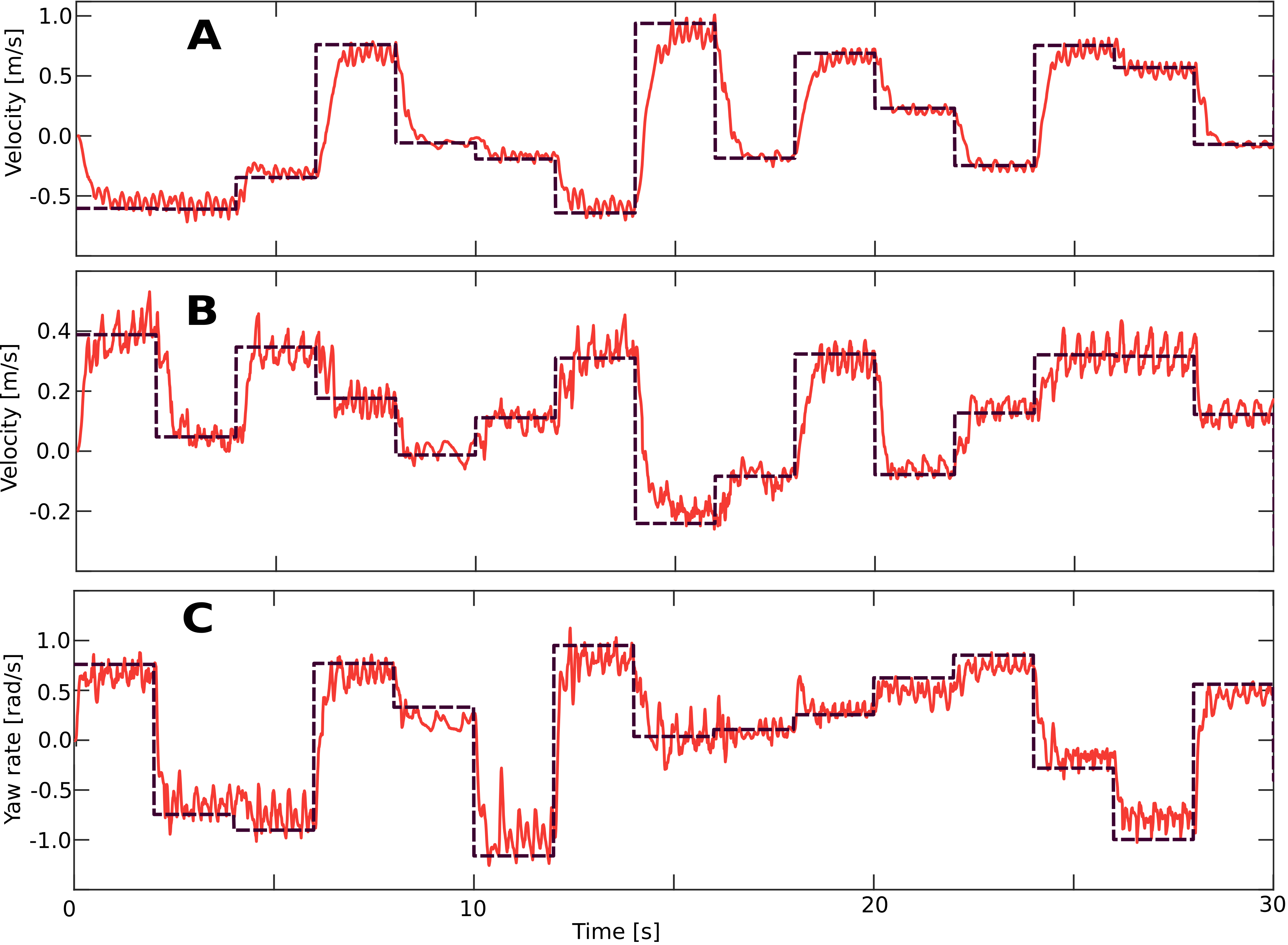}
    \caption{\textbf{Base velocity tracking performance of the learned controller while following random commands.} (\textbf{A}) Forward velocity, (\textbf{B}) Lateral velocity, (\textbf{C}) Yaw rate. For all graphs, the dotted lines represent the commanded velocity and the solid lines represent the measured velocity. All commands are followed with a reasonable accuracy even when the commands are given in a random fashion.}
    \label{fig:tracking_learned}
\end{sidewaysfigure*}

\begin{sidewaysfigure*}
\centering
    \includegraphics[width=0.8\textwidth]{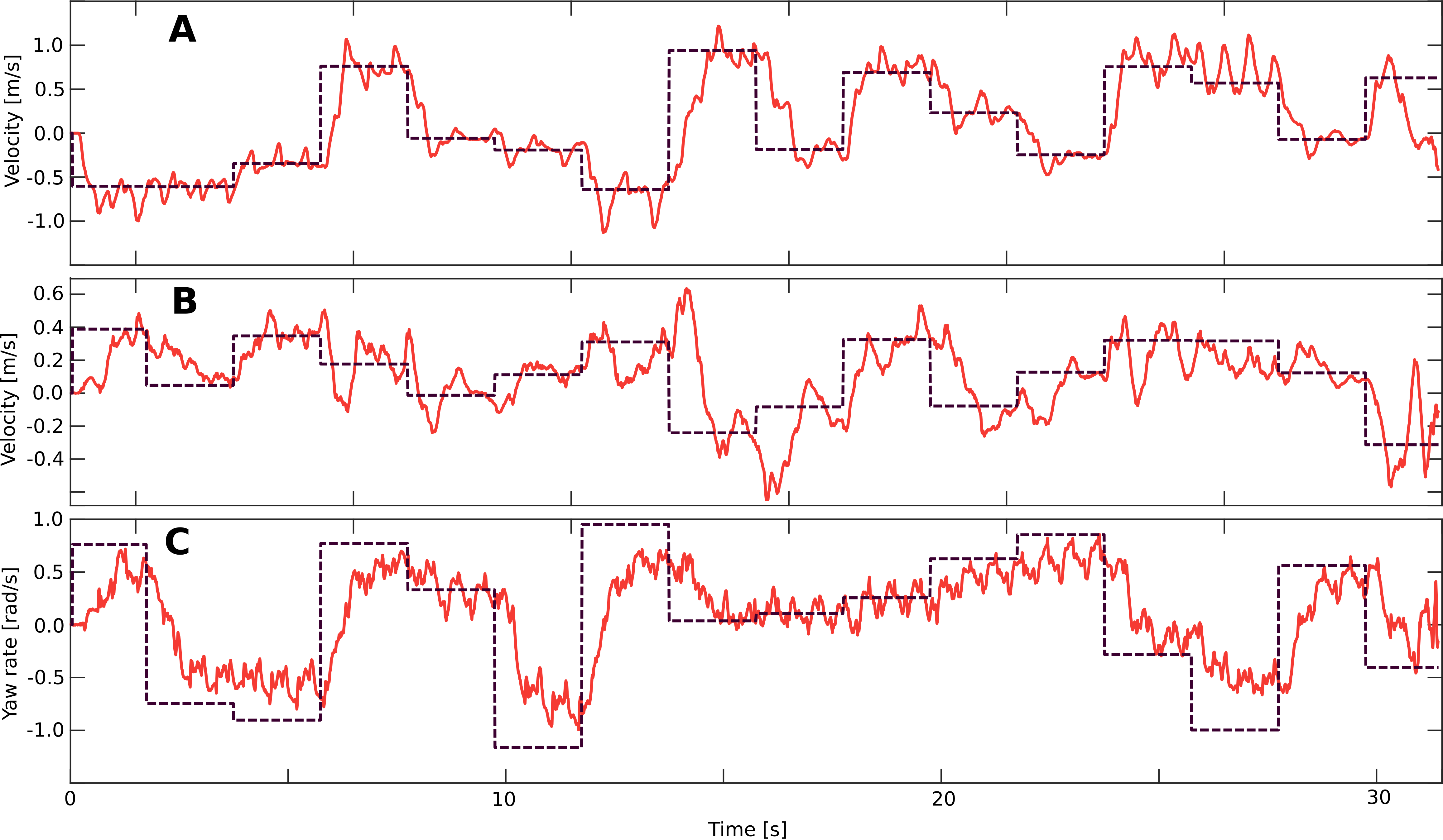}
    \caption{\textbf{Base velocity tracking performance of the best existing method while following random commands.} (\textbf{A}) Forward velocity, (\textbf{B}) Lateral velocity, (\textbf{C}) Yaw rate. For all graphs, the dotted lines represent the commanded velocity and the solid lines represent the measured velocity. The tracking performance is significantly worse than the learned policy.}
    \label{fig:tracking_modelbased}
\end{sidewaysfigure*}

\begin{figure*}
\centering
    \includegraphics[width=0.75\textwidth]{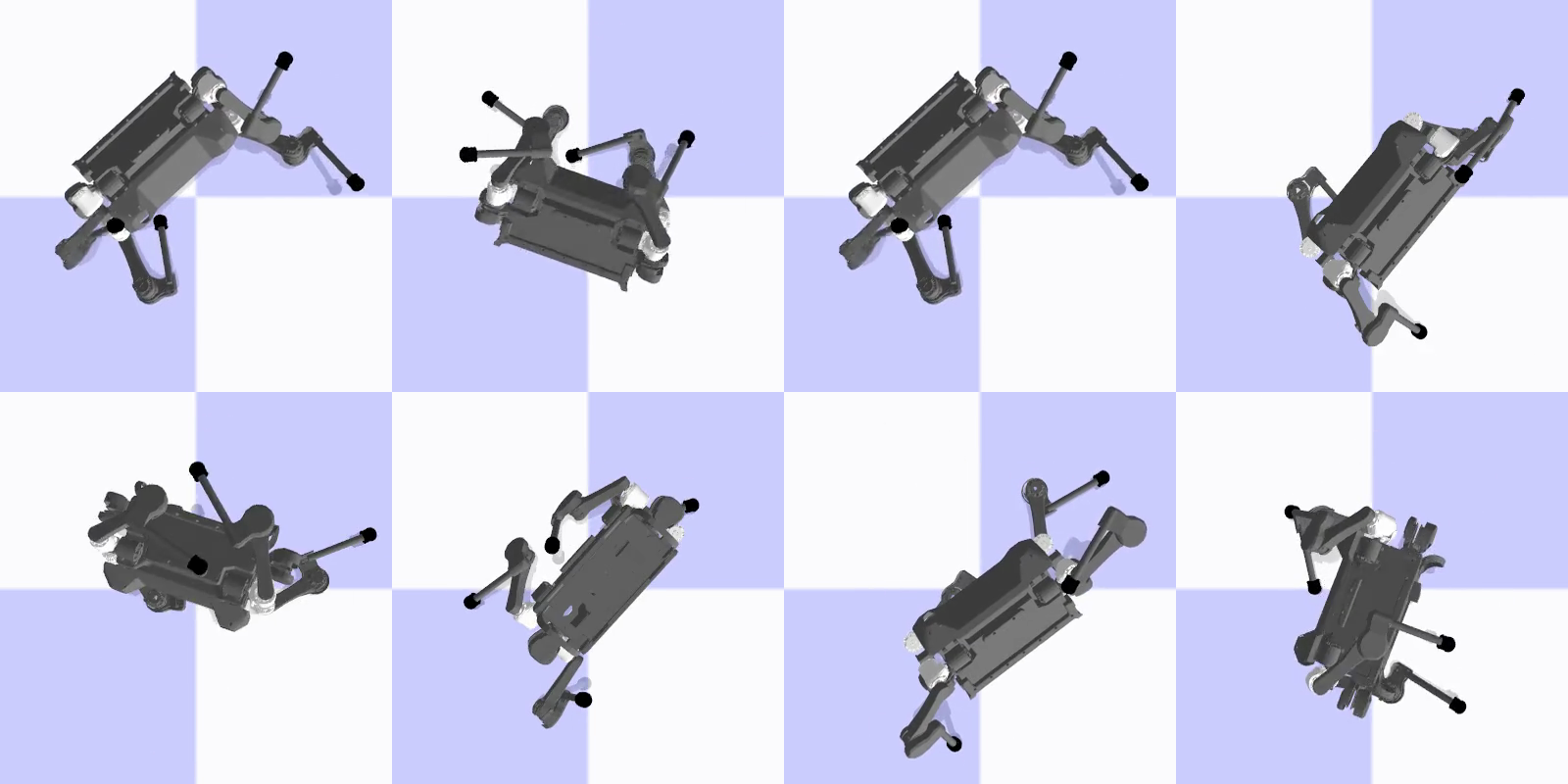}
    \caption{Sampled initial states for training a recovery controller}
    \label{fig:initial_states}
\end{figure*}
\end{document}